\documentclass[conference]{IEEEtran}
\IEEEoverridecommandlockouts

\usepackage{cite}
\usepackage{amsmath,amssymb,amsfonts}
\usepackage{graphicx}
\usepackage{textcomp}
\usepackage{xcolor}
\usepackage{booktabs}
\usepackage{float}
\usepackage{tabularx}
\usepackage{array}
\usepackage{url}
\usepackage[colorlinks=true, linkcolor=blue, urlcolor=blue, citecolor=blue]{hyperref}

\newcolumntype{Y}{>{\raggedright\arraybackslash}X}
\newcolumntype{C}{>{\centering\arraybackslash}X}

\def\BibTeX{{\rm B\kern-.05em{\sc i\kern-.025em b}\kern-.08em
    T\kern-.1667em\lower.7ex\hbox{E}\kern-.125emX}}
\usepackage{tcolorbox}
\tcbuselibrary{raster}

\begin{document}

\title{Tool-Augmented Large Language Models for Reliable Medical Calculation from Patient Notes}

\author{\IEEEauthorblockN{
Yang Zhong \textsuperscript{1},
Ling Luo\textsuperscript{1,*}, 
Xuanlyu Zhu\textsuperscript{1},
Yanhua Wang\textsuperscript{2}, %
Yuanyuan Sun\textsuperscript{1},
Jian Wang\textsuperscript{1},
Hongfei Lin\textsuperscript{1}
}
\IEEEauthorblockA{
\textsuperscript{1}College of Computer Science and Technology, Dalian University of Technology, Dalian, China \\
\textsuperscript{2}Air Force Communications NCO Academy, Dalian, China \\
\textsuperscript{*}To whom correspondence should be addressed: lingluo@dlut.edu.cn}
}

\maketitle

\begin{abstract}
Medical calculators translate raw patient observations into clinically actionable values and scores. However, applying these calculators directly to free-form patient notes requires precise variable extraction combined with the exact execution of complex formulas, scoring rules, and output conventions. Relying solely on direct large language model (LLM) prompting is unreliable in this context, as rule recall, parameter extraction, arithmetic reasoning, and formatting become entangled within a single generative step. To address these limitations, we propose a tool-augmented LLM framework for reliable medical calculation from patient notes. The framework retrieves and structures calculator specifications, generates reusable executable tools, and uses a constrained extractor to map each patient note to calculator-ready inputs. The tools then perform deterministic calculation, separating clinical language understanding from rule execution and arithmetic. Evaluated on the MedCalc-Bench dataset, our method achieves 86.73\% final-answer accuracy, outperforming the one-shot prompting and supervised fine tuning-only baselines by 24.0 and 17.27 percentage points. Ablation and error analyses show that these performance gains are primarily driven by rule-grounded tool synthesis and schema-constrained extraction, whereas remaining failures are concentrated in clinical-semantic extraction and benchmark-specific output contracts. By grounding calculator logic in executable tools, the proposed approach alleviates arithmetic hallucinations and improves the computational reliability of the final clinical outputs.
\end{abstract}


\section{Introduction}

Medical calculators convert patient observations into clinically meaningful values, risk scores, or treatment-relevant estimates. Clinicians use calculator resources such as MDCalc \cite{mdcalcweb} to estimate renal function, stratify cardiovascular and thromboembolic risk, grade disease severity, estimate gestational age, and support medication dosing. However, performing these calculations manually can be time-consuming and prone to error, particularly in settings with limited clinical capacity. The projected global shortage of health workers further underscores the need for reliable methods that can reduce routine computational burden while preserving transparency and clinical oversight \cite{world2016global}. Automating medical calculations from free-form patient notes aims to provide reliable computational support in settings where clinician time is limited.

This task differs fundamentally from conventional medical question answering. A system must identify every variable required by the target calculator, including numerical measurements, categorical variables, Boolean criteria, dates, time intervals, and units. These variables may be expressed indirectly, distributed across a note, negated, or implied by diagnoses, medications, and clinical history. The system must then apply the appropriate formula, scoring rule, threshold, boundary condition, and output format. A missed risk factor, an incorrectly interpreted time window, or an improperly normalized unit can change the final result. Reliable medical calculation therefore requires both accurate clinical information extraction and deterministic rule execution, together with a trace that makes errors inspectable.

Large language models (LLMs) are promising interfaces for this task because they can interpret unstructured clinical text, follow instructions, generate structured representations, and interact with external tools. Nevertheless, direct prompting remains unreliable for medical calculation \cite{khandekar2024medcalc}. When formula recall, parameter extraction, mathematical arithmetic, and output formatting are entangled within a single generative step, a hallucinated formula or a minor arithmetic error can silently corrupt the final answer without revealing the source of failure. Prior work has sought to address these limitations through retrieval-augmented prompting, supervised or reinforcement learning, program synthesis, and tool-using agents. In particular, tool-augmented systems can improve numerical reliability by delegating arithmetic to validated calculators \cite{jin2024agentmd,goodell2025large}. However, existing approaches commonly rely on fixed, manually implemented calculator libraries, which limits coverage to calculators already available as tools. In addition, many systems still combine parameter extraction and calculation within the same reasoning loop, making it difficult to distinguish errors in clinical understanding from errors in calculator implementation or output formatting.

Taiyi demonstrates bilingual biomedical multitask adaptation through supervised fine-tuning \cite{luo2024taiyi}. A recent survey reviews collaboration between LLMs, symbolic solvers, and verifiers for mathematical problem solving \cite{he2025mathsurvey}.

To address these limitations, we propose a tool-augmented method that separates clinical language understanding from deterministic medical calculation. Our method first retrieves evidence for a target calculator and structures its specification, including required inputs, formulas or scoring rules, boundary conditions, and expected outputs. It then synthesizes a reusable executable calculator tool and validates the tool on examples from the same calculator, using execution feedback to identify and repair implementation errors. At inference time, a schema-constrained extraction model maps each patient note to calculator-ready structured inputs, and the tool performs the final computation deterministically.

Our contributions are as follows.
\begin{itemize}
    \item We propose a tool-augmented framework for medical calculation from patient notes that combines rule-grounded retrieval, on-demand executable tool synthesis with validation and repair, schema-constrained parameter extraction, and deterministic tool execution. This decoupled architecture effectively mitigates formula hallucinations and arithmetic errors, and supports tool construction beyond predefined tool libraries.
    \item We establish a standardized evaluation on the corrected MedCalc-Bench benchmark using a unified data split and official metrics. We compare prompting baselines, multiple tool-writing systems, and extraction models under the same evaluation setting, avoiding comparisons aggregated across different benchmark versions.
    \item Experimental results show that the best configuration achieves 86.73\% final-answer accuracy and a 97.47\% parameter matching rate on 1,100 test cases, outperforming one-shot prompting baselines by 24.0 percentage points. Tool-level evaluation, extraction-model comparisons, category-level analyses, ablations, and case studies further characterize the sources of improvement and remaining errors.
\end{itemize}

The code for our framework is available online.\footnote{\url{https://github.com/benjiesun/tool-augmented-medical-calculation}}

\section{Related Work}

\subsection{Medical Calculation Benchmarks}
MedCalc-Bench \cite{khandekar2024medcalc} established medical calculation as an evaluation target distinct from knowledge-oriented medical question answering. Unlike MedQA \cite{jin2021disease}, PubMedQA \cite{jin2019pubmedqa}, MedMCQA \cite{pal2022medmcqa}, Med-PaLM \cite{singhal2023large}, and HealthBench \cite{arora2025healthbench}, a medical calculator is defined by an explicit equation, a scoring rule, or branch logic. Consequently, a model needs to accurately extract the required patient variables from free text and then apply that specific rule deterministically. This task therefore jointly evaluates structured clinical extraction and exact execution rather than medical knowledge alone. The MedCalc-Bench baselines still err frequently even when the target calculator is given, exposing a persistent gap between recalling a formula and applying it to a specific note \cite{khandekar2024medcalc}. The original MedCalc-Bench results show that LLMs still fail frequently even when the target calculator is given, with errors arising from entity extraction, rule selection, and arithmetic execution \cite{khandekar2024medcalc}. Subsequent audits further show that many original labels are erroneous, ambiguous, or incomputable, and that open-book access to calculator rules changes the interpretation of model performance \cite{ye2025scalable,krohn2026medcalc}. These findings motivate our use of the corrected MedCalc-Bench release as the primary benchmark and our emphasis on evaluation within one fixed dataset version and answer contract.

\subsection{LLMs for Medical Calculation}
Existing approaches for LLM-based medical calculation can be grouped into three main lines: retrieval- and prompting-based methods, supervised training-based methods, and tool-calling methods \cite{wei2022chain,wang2022self,lewis2020retrieval,gao2023pal,chen2022program,yao2022react,schick2023toolformer,qin2024tool}. First, retrieval and prompting methods keep the model frozen and restructure the input. MedRaC \cite{wang2025scores}, for example, decomposes medical calculation into formula selection, entity extraction, calculation, and final-answer checking, and argues that final-answer accuracy alone cannot localize failures; however, its retrieval and Python-execution steps are still performed within the per-case inference pipeline, so formula-selection, code-execution, and answer-formatting errors can still propagate to the final answer. %
Second, supervised training-based methods encode calculator knowledge in model weights. MedCalc-R1 \cite{wang2025medcalcr} combines supervised fine-tuning with a knowledge-guided GRPO reward, and MedCalc-Eval/MedCalc-Env build training data and an interactive environment \cite{mao2025medcalc}. These methods can improve accuracy, but calculator updates, revised rules, or changed answer contracts may require additional training. Third, tool-calling methods delegate computation to external calculators or tool libraries. Prior work has shown that tool use can substantially reduce numerical errors in clinical calculation settings \cite{goodell2025large,jin2024agentmd}. Related works such as RiskAgent \cite{liu2025riskagent}, MeNTi \cite{zhu2025menti}, and MedMCP-Calc \cite{zhu2026medmcp} connect agents to calculators through retrieval, nested tool calling, and Model Context Protocol integration. These approaches improve numerical stability, but they often assume that the target calculator already exists as a manually implemented tool. Many also keep parameter extraction and tool invocation within the same agent loop, which makes it difficult to separate errors in clinical interpretation from errors in tool selection, tool implementation, or output formatting.

Our method differs from these methods in two respects. First, it does not assume a fixed, hand-built calculator library. Instead, it synthesizes an executable calculator tool from retrieved rule evidence and improves the tool through execution-based validation and repair. Second, it separates schema-constrained parameter extraction from deterministic tool execution. This design reduces the coupling between clinical understanding and computation, while producing an auditable trace of the retrieved rule evidence, generated tool, extracted parameters, tool input, and tool output. As a result, errors can be localized more clearly to rule retrieval, tool construction, parameter extraction, or benchmark-specific output handling.

\section{Method}

\begin{figure*}[!t]
\centering
\includegraphics[width=0.95\textwidth]{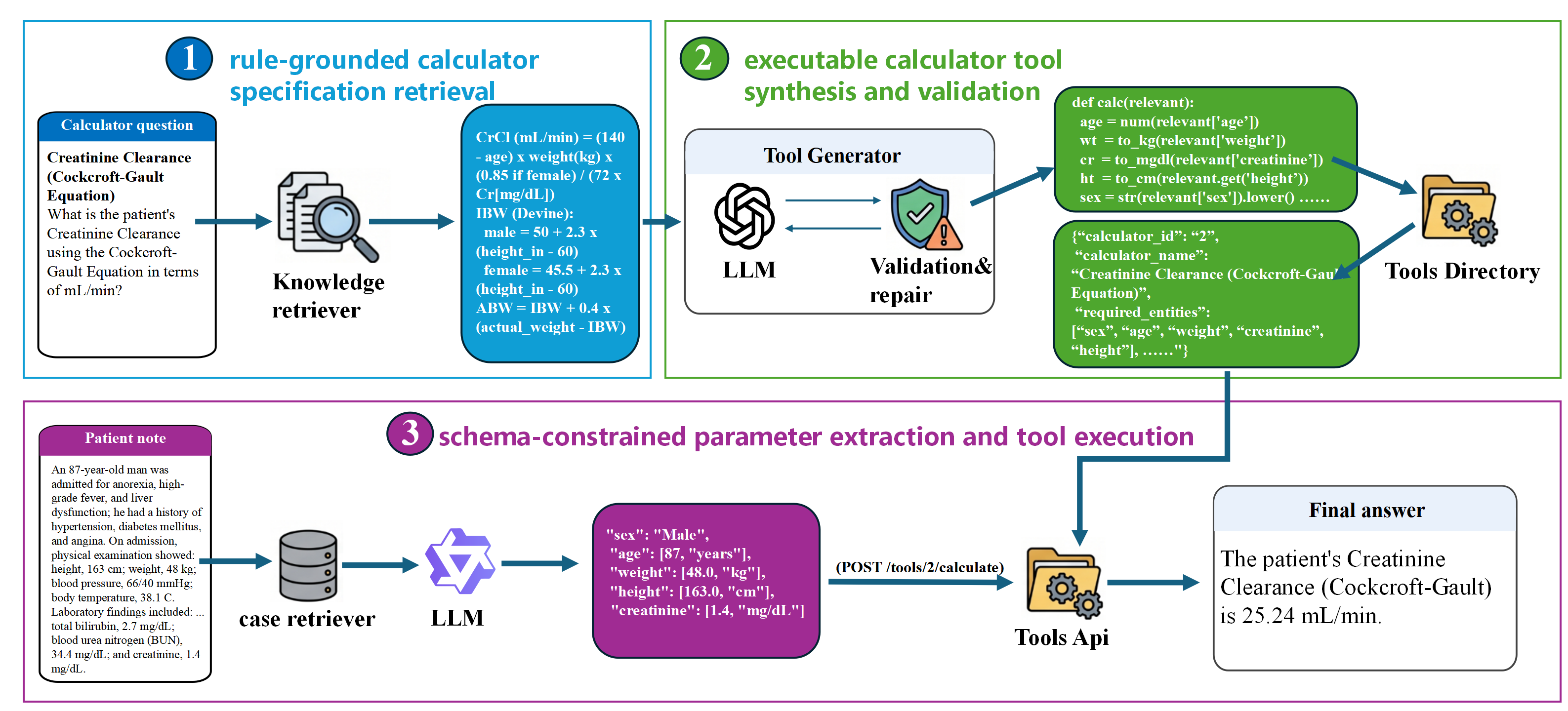}
\caption{Overview of the proposed method. It is illustrated with the Cockcroft–Gault creatinine-clearance calculator. The workflow retrieves and structures calculator rules, synthesizes and validates an executable calculator tool, extracts calculator inputs from a patient note, and executes the validated tool deterministically.}
\label{fig:framework}
\end{figure*}

Our approach formulates medical calculation from patient notes as a tool-augmented workflow with three main steps shown in Figure~\ref{fig:framework}: (1) rule-grounded calculator specification retrieval, (2) executable calculator tool synthesis and validation, and (3) schema-constrained parameter extraction and tool execution. Next, we will first describe the task definition and then detail each step.

\subsection{Task Definition}
Let an input be denoted by $x=(n,q)$, where $n$ is a patient note, $q$ is the calculation question. The goal is to produce an answer $y$ that is consistent with the reference calculator $c$ and the benchmark-specific output contract. Our workflow can be expressed as:
\begin{align}
    (c, r_c, S_c) &= \mathrm{RetrieveAndStructure}(q),\\
    \tau_c &= \mathrm{GenerateTool}(c,r_c,S_c),\\
    z &= \mathrm{Extract}(n,q,c,S_c,K_c),\\
    y &= \mathrm{Execute}(\tau_c,z)
\end{align}
where $r_c$ is the rule specification for calculator $c$, $S_c$ is the calculator parameter schema, $\tau_c$ is the generated executable tool, $K_c$ is a set of same-calculator demonstrations, and $z$ is the structured parameter object extracted from the patient note. The LLM is responsible for organizing calculator rules, generating tools, and interpreting clinical language, whereas the validated executable tool performs the final medical calculation.

\subsection{Rule-Grounded Calculator Specification Retrieval}
The first stage retrieves and structures the specification of the target calculator. First, based on the input question $q$, the system searches medical calculation resource websites such as MDCalc to find the corresponding target calculator $c$. When a calculator specification has already been stored in memory, the workflow reuses the cached version. Otherwise, the knowledge retriever searches for calculator evidence together with benchmark-provided metadata where available. A rule-curation LLM converts the retrieved evidence into a structured calculator specification (Figure~\ref{fig:prompts} (1) shows the specific prompt):
\begin{equation}
    r_c = \{\mathrm{name}, \mathrm{inputs}, \mathrm{formula}, \mathrm{conditions}, \mathrm{output}\}.
\end{equation}
The corresponding calculator parameter schema $S_c$ specifies the complete set of calculator inputs, their value types, allowed units, and permitted categorical values. The structured specification is stored with provenance information so that later cases involving the same calculator can reuse the retrieved evidence without repeated search.

\begin{figure*}[!t]
\centering
\begingroup

\definecolor{PromptBlue}{HTML}{426781}
\definecolor{PromptGreen}{HTML}{527565}
\definecolor{PromptPurple}{HTML}{746187}

\newcommand{\PromptItem}[1]{%
  \par\noindent
  \hangindent=8pt
  \hangafter=1
  \makebox[8pt][l]{%
    \textcolor{black!55}{\textbullet}%
  }%
  #1\par\vspace{2pt}%
}

\begin{tcbraster}[
  raster columns=3,
  raster width=\textwidth,
  raster column skip=7pt,
  raster equal height=rows,
  raster valign=top,
  colback=white,
  coltext=black!90,
  coltitle=black!90,
  boxrule=0.45pt,
  arc=1.2mm,
  boxsep=0pt,
  left=6pt,
  right=6pt,
  top=6pt,
  bottom=5pt,
  toptitle=5pt,
  bottomtitle=5pt,
  fonttitle=\fontfamily{ppl}\fontsize{9}{11}\selectfont\bfseries,
  fontupper=\fontfamily{ppl}\fontsize{8.5}{10.5}\selectfont,
  before upper={
    \raggedright
    \setlength{\parindent}{0pt}
    \setlength{\parskip}{0pt}
  },
  valign=top,
  before skip=0pt,
  after skip=0pt
]

\begin{tcolorbox}[
  title={(1) Rule Curation Prompt},
  colframe=PromptBlue,
  colbacktitle=PromptBlue!9!white
]
You are a medical calculator rule curator.
Read web evidence and output STRICT JSON only.

\par\vspace{5pt}

\PromptItem{Calculator Name \& Question.}

\PromptItem{Retrieved evidence snippets and source hits.}

\PromptItem{Return calculator rule: name, URL,
formula or scoring rule, input variables, conditions,
boundary cases, output format, evidence text.}

\PromptItem{Return calculator schema:
required\_fields; value kinds; allowed units;
categorical enums; nullable/default behavior.}

\PromptItem{Do not invent facts absent from evidence.}
\end{tcolorbox}
\begin{tcolorbox}[
  title={(2) Tool Synthesis Prompt},
  colframe=PromptGreen,
  colbacktitle=PromptGreen!9!white
]
You are an expert medical calculator tool builder.
Output STRICT JSON only, no markdown, no explanation.

\par\vspace{5pt}

\PromptItem{Inputs: calculator name;
expected output type; curated rule; schema $S_c$;
training examples; RAG rule references.}

\PromptItem{Write robust Python: def calc(relevant):
values may be scalar, [value, unit], tuple, or dict.}

\PromptItem{Add safe parsing helpers for numbers,
text, Booleans, dates, and units.}

\PromptItem{Implement formula/scoring logic,
branches, boundary cases, rounding,
and output contract.}

\PromptItem{If required inputs are missing,
return the benchmark-specified missing-value output.}

\PromptItem{Deterministic, runnable,
no external dependencies.}

\PromptItem{Return JSON:
\{"required\_entities": [...],
"tool\_code": "def calc(relevant): ..."\}}
\end{tcolorbox}
\begin{tcolorbox}[
  title={(3) Extraction Prompt},
  colframe=PromptPurple,
  colbacktitle=PromptPurple!9!white
]
You are extracting parameters for a medical calculator.
Return only one valid JSON object, no markdown
or extra keys.

\par\vspace{5pt}

\PromptItem{Calculator ID / Name; Question.}

\PromptItem{Required parameters from schema.}

\PromptItem{JSON schema for the required output:
value kinds, allowed units, enums, nullable fields.}

\PromptItem{Field specs and $k=3$
same-calculator examples.}

\PromptItem{Return measured values as [value, unit];
preserve allowed source units when possible.}

\PromptItem{Use null for unsupported required fields;
use false for absent Boolean criteria.}

\PromptItem{For categorical fields,
explicit negative evidence maps to
the negative enum value.}

\PromptItem{Patient note $n$.}

\PromptItem{Output JSON $z$ only.}
\end{tcolorbox}

\end{tcbraster}
\endgroup

\caption{Condensed prompts used for calculator rule
curation, executable tool synthesis, and
schema-constrained parameter extraction.}
\label{fig:prompts}
\end{figure*}

\subsection{Executable Calculator Tool Synthesis and Validation}
The second stage transforms the retrieved calculator specification into reusable executable code. Medical calculators may involve numerical formulas, categorical scoring rules, Boolean criteria, conditional branches, unit conversions, boundary conditions, rounding rules, and calculator-specific output conventions. Directly asking an LLM to recall and apply all of these requirements during answer generation can lead to incorrect formulas, omitted conditions, arithmetic errors, or format mismatches. We therefore convert the rule into Python code before test-time inference. 

Given the structured rule specification $r_c$ and parameter schema $S_c$, the tool-writing LLM generates a Python calculator function \texttt{calc(relevant)}. The function validates and parses inputs, implements the required formula or scoring logic, handles relevant branches and boundary cases, and returns an output consistent with the calculator’s expected contract. Figure~\ref{fig:prompts} (2) shows the tool-builder prompt.

Tool generation is iterative. Each candidate tool is evaluated on eight training examples associated with the same calculator. Runtime failures, missing-field errors, and output mismatches are returned to the tool-writing model together with the expected and observed results. The model may revise the implementation for up to four validation-and-repair rounds. Once validated, the tool is cached and reused for subsequent cases involving the same calculator. This design reduces repeated tool-generation calls and promotes consistent behavior across patient notes.

Validation samples are grouped by calculator ID and eight are randomly drawn without replacement from the historical training set for each calculator. They are not selected by difficulty or field coverage. All repair rounds reuse the same eight samples, executing each tool with annotated parameters and comparing its outputs with reference answers. This sampling differs from the coverage-based selection of the three extraction demonstrations. Eight random checks do not guarantee branch or boundary coverage, and deterministic execution does not guarantee a correct tool implementation.

\subsection{Schema-Constrained Parameter Extraction and Tool Execution}
The third stage operates on individual patient notes. The calculator tools are generated once per calculator and then reused, whereas the repeated inference-time task is parameter extraction. Therefore, we fine-tuned a smaller Qwen3-8B model for this high-frequency stage, while reserving the tool-writing LLM for calculator construction and validation. This division can reduce computational cost and enable the extraction component to be deployed locally when required by data-governance constraints. 

Before extraction, a case retriever selects $k=3$ same-calculator examples, denoted by $K_c$, and provides them together with the calculator parameter schema $S_c$. Calculator-specific schemas are necessary because medical calculation requires more than generic entity recognition. Depending on the calculator, the required inputs may include continuous measurements, Boolean risk factors, disease categories, dates, gestational ages, time intervals, or values paired with units.

The extraction model receives the patient note $n$, the question $q$, the calculator $c$, the examples $K_c$, and the schema $S_c$. Figure~\ref{fig:prompts} (3) shows the extraction task prompt. The constrained JSON object is returned:
\begin{equation}
    z = \{s_1:v_1,\ldots,s_m:v_m\},
\end{equation}
where each value may be a number, Boolean value, categorical label, date, time interval, or value-unit pair. The structured parameter object is then sent to the tool-serving interface, which loads the cached tool $\tau_c$, executes the calculator using \texttt{calc(z)}, and returns the final calculation result. The LLM does not perform the final formula application or arithmetic at this stage; these operations are executed deterministically by the executable tool. 

We derive schemas from calculator specifications rather than from individual annotated examples. Across the 55 calculators in the benchmark, the schemas contain 328 required fields. For few-shot extraction, we begin with the public exemplar for each calculator and select two additional training examples greedily to cover required fields not represented in the initial example. This procedure yields three demonstrations per calculator and covers all 328 required fields without using test cases as demonstrations.

Memory serves as an auxiliary mechanism rather than the main methodological contribution. It stores calculator specifications, schemas, generated tools, selected examples, execution traces, and historical errors. This supports reuse, reduces repeated retrieval, and improves auditability.

\section{Experiments}

\subsection{Dataset and Experimental Settings}
We conduct controlled experiments on MedCalc-Bench Verified v1.0.7 \cite{medcalcverifiedweb}, a corrected version of MedCalc-Bench that includes revised labels, answer contracts, calculator specifications, and executable reference implementations for 55 medical calculators. The dataset contains 10,538 training instances and 1,100 test instances. We do not use the original MedCalc-Bench v1.0 for evaluation because physician-supervised audits have identified erroneous, ambiguous, and incomputable cases in the original benchmark \cite{ye2025scalable,krohn2026medcalc}.

We report three evaluation metrics for different modules. Final-answer accuracy is the proportion of test cases judged correct by the official benchmark evaluator. Parameter matching rate (PMR) is the proportion of required calculator parameters for which the extracted value matches the gold annotation. Tool-level accuracy evaluates the generated calculator tools using gold parameters, thereby isolating calculator implementation quality from parameter-extraction quality.

Unless otherwise specified, GPT-5.5 is used to generate calculator tools, and Qwen3-8B \cite{yang2025qwen3} is used for parameter extraction. Table~\ref{tab:sft} summarizes the main configuration for supervised fine-tuning of the extraction models.

\begin{table}[!htbp]
\caption{Main supervised fine-tuning configuration for the parameter-extraction model.}
\label{tab:sft}
\centering
\footnotesize
\begin{tabularx}{\columnwidth}{@{}lY@{}}
\toprule
Item & Setting\\
\midrule
Base models &  Qwen3-8B\\
Tuning method & LoRA adapter on frozen base model\\
LoRA rank / alpha / dropout & 16 / 32 / 0.05 \\
Learning rate  & $5{\times}10^{-5}$ \\
Epochs & 5 \\
global batch size & 8 \\
Max sequence length & 8,192 \\
\bottomrule
\end{tabularx}
\end{table}

\subsection{Main Results}
Table~\ref{tab:main} presents the main results on the MedCalc-Bench Verified test set. The baseline results show that direct SFT and prompting-based reasoning strategies provide limited improvements when clinical parameter extraction, calculator-rule application, arithmetic, and output formatting must be handled within a single generation. Our method achieves 86.73\% final-answer accuracy, outperforming the strongest one-shot prompting baseline and the SFT-only baseline based on the same Qwen3-8B backbone by 24.00 and 17.27 percentage points, respectively. These results indicate that decomposing the task and delegating rule execution and arithmetic to medical calculator tools substantially improves end-to-end performance.

The table also includes previously reported results as references. %
MedRaC is a training-free Formula-RAG and Python-execution pipeline \cite{wang2025scores}. The original paper reports an overall accuracy of 64.68\% in its ablation section. Based on the surrounding discussion, this result follows the LLaMA3.1-8B-Instruct analysis, and it is evaluated on the authors' cleaned MedCalc-Bench set rather than the full MedCalc-Bench Verified v1.0.7 test set. The open-book rows provide calculator specifications at inference time, with the guided variant adding task-specific instructions \cite{krohn2026medcalc}. Claude Sonnet 4.6 is reported under a code-augmented agent loop that uses a Python REPL for computation, averaged over five runs on the latest verified dataset \cite{anthropic2026claudesonnet46systemcard}.
Compared to these methods, our approach achieves competitive performance. Notably, some reported results were evaluated on a subset rather than the full 1,100-case Verified test split. Our workflow delegates calculator-rule interpretation, branch logic, rounding behavior, and output conventions to a reusable, validated tool. Consequently, the extraction model is responsible only for mapping the patient note to the required calculator schema. This separation prevents the need for a single generative step to be simultaneously clinically accurate, mathematically exact, and compliant with complex output conventions.

The external rows in Table II are contextual comparisons, whereas the rows from Direct prompting through Ours use our controlled evaluation. The 0.49-percentage-point difference from the externally reported 86.24\% does not establish a statistically significant advantage.

\begin{table}[!htbp]
	\caption{Main results on MedCalc-Bench Verified v1.0.7.}
	\label{tab:main}
	\centering
	\scriptsize
	\resizebox{\columnwidth}{!}{%
		\begin{tabularx}{\columnwidth}{@{}lYc@{}}
			\toprule
			Method & Model & Score (\%)\\
			\midrule
			MedRaC\textsuperscript{*} \cite{wang2025scores}
			& LLaMA3.1-8B-Instruct & 64.68\\
			Open-book guided prompting \cite{krohn2026medcalc}
			& GLM-4.6V & 82.00\\
			Open-book prompting\textsuperscript{*} \cite{krohn2026medcalc}
			& GLM-4.7 & 85.50\\
			Code agent (Python REPL) \cite{anthropic2026claudesonnet46systemcard}
			& Claude Sonnet 4.6 & 86.24\\
			\midrule
			Direct prompting & Qwen3-8B & 14.00\\
			Zero-shot CoT & Qwen3-8B & 31.55\\
			One-shot prompting & Qwen3-8B & 62.73\\
			One-shot prompting & DeepSeek-R1-Distill-Qwen-7B & 55.45\\
			One-shot prompting & Llama-3.1-8B & 38.36\\
			SFT-only & Qwen3-8B & 69.46\\
			Ours & GPT-5.5 tool generation + Qwen3-8B-SFT extraction & \textbf{86.73}\\
			\bottomrule
			\addlinespace[2pt]
			\multicolumn{3}{@{}p{\columnwidth}@{}}{\scriptsize \textit{Note:} \textsuperscript{*}denotes the results reported on the sub test set.}
		\end{tabularx}
	}
\end{table}

\subsection{Effect of Generated Tool Quality on Medical Calculation Performance}
Table~\ref{tab:tools} evaluates the effect of generated calculator-tool quality on final medical-calculation performance. Tool-level accuracy is measured by providing each generated tool with gold parameters. The final-answer accuracy then evaluates the same tool set with a fixed Qwen3-8B-SFT extractor. The results show that deterministic execution improves reliability when the generated calculator tool faithfully implements the relevant calculator rule. GPT-5.5 tools generated with rule-grounded retrieval and validation-and-repair achieve the highest tool-level accuracy and final-answer accuracy. Compared with tools generated by Qwen3.5-Plus, the GPT-5.5 tools show a clear performance advantage and produce no runtime errors. Moreover, removing web retrieval and iterative repair reduces both tool-level accuracy and final-answer accuracy by more than 10 percentage points, despite using the same extraction model.

With GPT-5.5 and the Qwen3-8B-SFT extractor fixed, retrieval plus repair raises final accuracy from 73.27\% to 86.73\% (+13.46 percentage points), supporting their joint contribution beyond the tool-writing model alone. The comparison does not separate retrieval from repair.

\begin{table}[!htbp]
\caption{Generated Tool Quality and End-to-End Calculation Performance.}
\label{tab:tools}
\centering
\scriptsize
\begin{tabularx}{\columnwidth}{@{}Ycccc@{}}
\toprule
Tool setting & Model & Runtime  & Tool acc. & Final acc.\\
 & & errors &(\%) &  (\%)\\
\midrule
Direct generation  & GPT-5.5 & 0 & 74.45 & 73.27\\
Retrieval + repair & Qwen3.5-Plus & 21  & 84.45 & 76.18\\
Retrieval + repair & GPT-5.3-Codex & 4 & 89.00  & 82.82\\
Retrieval + repair  &  GPT-5.5 & 0  & 94.91  & 86.73\\
\bottomrule
\addlinespace[2pt]
\multicolumn{5}{@{}l@{}}{\scriptsize \textit{Note:} The last column pairs each tool set with the same Qwen3-8B-SFT extractor.}
\end{tabularx}
\end{table}

\subsection{Parameter-Extraction Model Comparison}

\begin{figure}[!htbp]
\centering
\includegraphics[width=0.96\columnwidth]{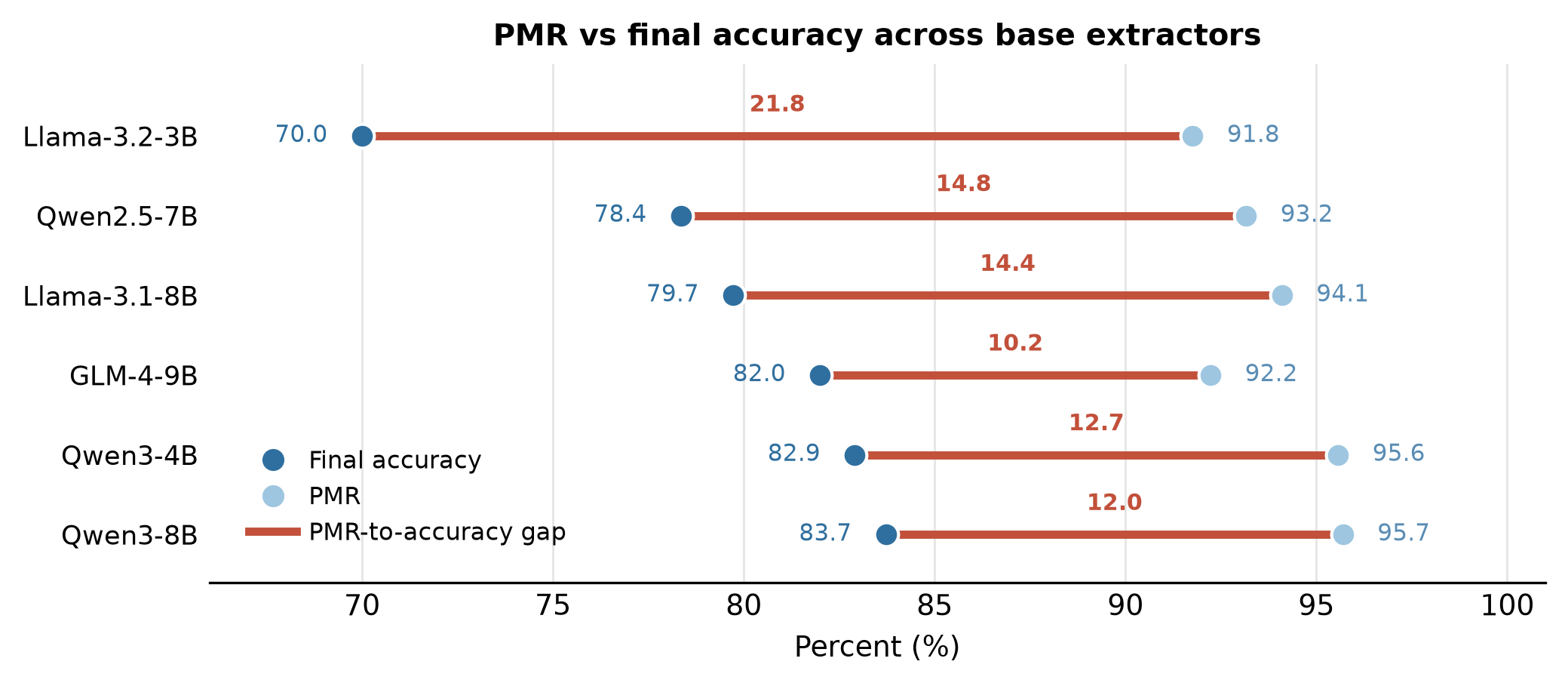}
\caption{Parameter matching rate and final-answer accuracy across base extractor families with a fixed GPT-5.5-generated tool set.}
\label{fig:extractors}
\end{figure}

Figure~\ref{fig:extractors} compares parameter-extraction models while holding the GPT-5.5-generated tool set fixed. The evaluated extractors include models from the Qwen3, Qwen2.5, Llama, and GLM families \cite{yang2025qwen3,qwen2.5,grattafiori2024llama,glm2024chatglm}. Once formula execution and arithmetic are delegated to tools, parameter extraction becomes the primary language-understanding challenge. The extractor must determine which clinical facts correspond to calculator fields and serialize them according to the required schema.

Across the evaluated models, higher PMR is generally associated with higher final-answer accuracy. The Qwen3 model achieved the best performance. However, while GLM-4 demonstrated a high Parameter Matching Rate (PMR), its final answer accuracy did not improve. This discrepancy arises because the two metrics measure performance at different levels. PMR is calculated over individual parameters, whereas final-answer accuracy is calculated at the case level. A case may have a high proportion of correctly extracted parameters but still receive an incorrect final answer when one clinically decisive field is missed or misclassified.

This discrepancy is especially important for calculators with Boolean risk factors, categorical inputs, date boundaries, or unit-sensitive measurements. For example, a missed comorbidity, an incorrectly inferred medication-related condition, or an incorrect date interpretation can alter the final score even when most other fields are correct. The case analysis presented later illustrates this issue using the HEART Score, where failure to infer hypercholesterolemia from atorvastatin use changes the final score despite otherwise accurate extraction.

\subsection{Category-Level Analysis}

\begin{figure}[!htbp]
\centering
\includegraphics[width=0.96\columnwidth]{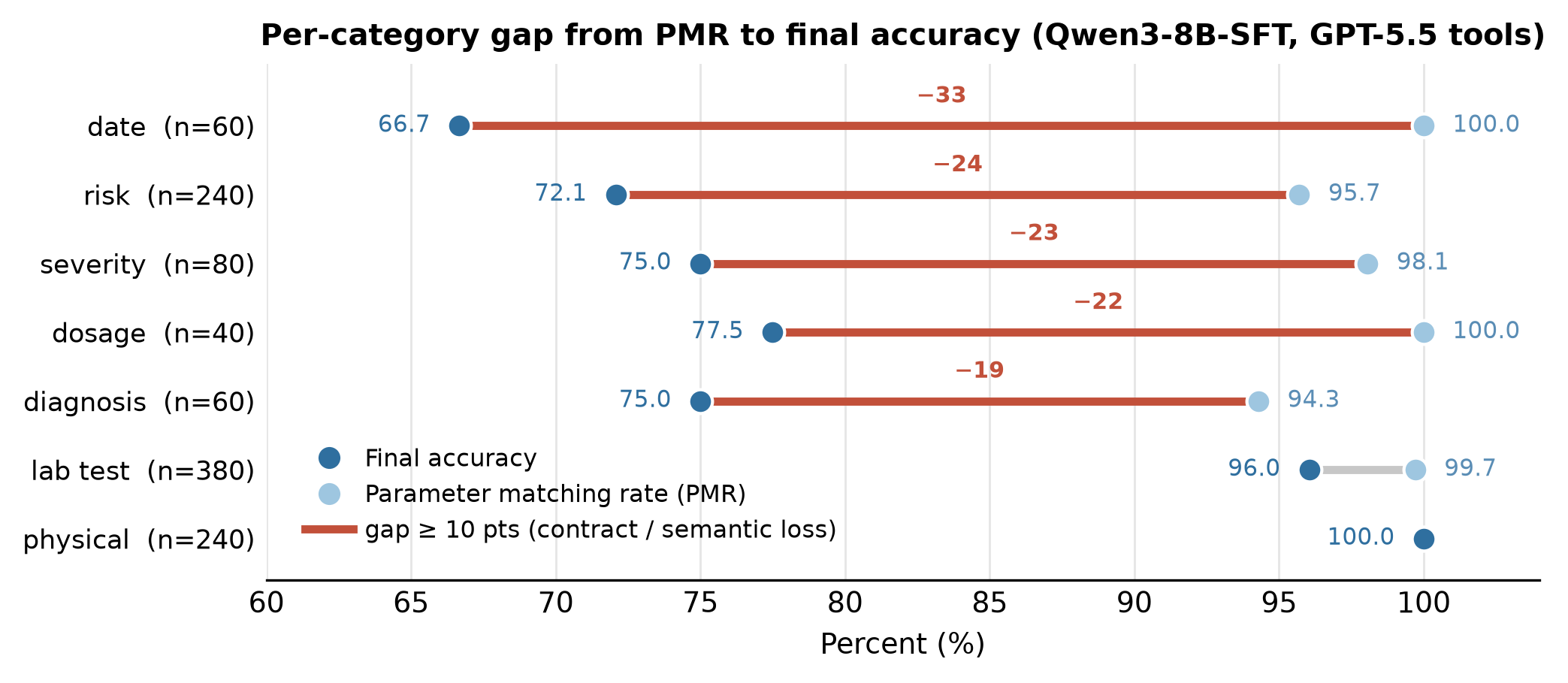}
\caption{Parameter matching rate and final-answer accuracy by calculator category for the best configuration.}
\label{fig:category}
\end{figure}

Figure~\ref{fig:category} compares PMR and final-answer accuracy across calculator categories for the best configuration. 

Low PMR primarily reflects errors in clinical interpretation and parameter extraction, whereas a large gap between PMR and final-answer accuracy motivates case-level inspection of possible errors after extraction, including boundary handling, unit conversion, rounding, or benchmark-specific output conventions. Physical-measurement and laboratory-test calculators achieve the strongest performance because their inputs are often explicit measurements and their rules are relatively direct. Once the correct values are extracted, deterministic execution resolves most of the remaining computation. Date and dosage calculators exhibit a different failure profile. Their input variables are often recoverable, but final correctness remains sensitive to date boundaries, unit conversions, rounding requirements, and output formatting. The date category shows the clearest discrepancy between PMR and final-answer accuracy, requiring case-level inspection to distinguish calculator or benchmark output-contract errors from failures to interpret the patient note. Risk, diagnosis, and severity calculators present the complementary challenge. These calculators often require clinical-semantic judgments involving implied comorbidities, symptom timing, disease categories, negation, and historical context. In such cases, a single Boolean or categorical error can change the final score.

PMR is parameter-level, whereas final-answer accuracy is case-level; their difference alone cannot localize failures to the generated tool or output contract.

\subsection{Ablation Analysis}
All ablation experiments use the same corrected test set and the official benchmark evaluator. The \textit{full workflow} combines GPT-5.5-generated tools with Qwen3-8B-SFT parameter extraction. The \textit{No extractor SFT} condition uses the same GPT-5.5-generated tools but replaces the fine-tuned extractor with the original Qwen3-8B model. The \textit{No schema/few-shot} condition removes the calculator schema and calculator demonstrations from the extraction prompt. The \textit{No tool} condition removes the tool-augmented execution and relies on the fine-tuned Qwen3-8B model to generate the final answer directly.

\begin{table}[!htbp]
    \caption{Component ablation results.}
    \label{tab:ablation}
    \centering
    \footnotesize
    \begin{tabularx}{\columnwidth}{@{}Ycc@{}}
        \toprule
        Variant & Final acc. (\%) & $\Delta$ \\
        \midrule
        Full workflow & 86.73 & -\\
        No extractor SFT & 83.73 & -3.00\\
        No schema/few-shot & 79.18 & -7.55\\
        No tool & 69.46 & -17.27\\
        \bottomrule
    \end{tabularx}
\end{table}

The results in Table~\ref{tab:ablation} show that both deterministic tool execution and constrained parameter extraction contribute substantially to final performance. First, the No tool condition results in the largest performance degradation (-17.27 percentage points), confirming that even a highly specialized fine-tuned model cannot reliably replicate the deterministic accuracy of an executable calculator. Second, the No schema/few-shot variant shows a substantial drop of 7.55 percentage points, highlighting that calculator-specific schemas (e.g., field definitions, unit requirements, and Boolean defaults) are fundamental to grounding the extraction process. Finally, the No extractor SFT condition results in a moderate decrease. This suggests that while inference-time schema constraints provide the necessary structure for the task, supervised fine-tuning helps the model handle complex clinical conventions, such as handling missing data, interpreting implied conditions, and ensuring consistent categorical labeling.

\subsection{Construction Usage and Inference Time}
Table~\ref{tab:construction_usage} summarizes the model calls and token consumption required to construct the 55 calculator tools using retrieval and iterative repair. Table~\ref{tab:inference_time} compares inference efficiency across four tool configurations on the 1,100-case test set using two NVIDIA RTX 5090 GPUs, with Qwen3-8B-SFT held fixed as the parameter extractor. We report the total inference time and the average time per case, computed by dividing the total runtime by the number of test cases. These measurements exclude tool construction and characterize average processing efficiency over the full test set rather than individual-request latency.

\begin{table}[H]\centering\footnotesize
\caption{Construction calls and token usage.}\label{tab:construction_usage}
\begin{tabularx}{\columnwidth}{@{}Yrrrr@{}}\toprule
Builder & Calls & Input & Output & Total\\\midrule
Qwen3.5-Plus & 144 & 243,571 & 147,350 & 390,921\\
GPT-5.3-Codex & 159 & 272,884 & 124,773 & 397,657\\
GPT-5.5 & 147 & 227,454 & 371,758 & 599,212\\
\bottomrule\end{tabularx}\end{table}
\begin{table}[H]\centering\footnotesize
\caption{Inference time with the same SFT extractor.}\label{tab:inference_time}
\begin{tabularx}{\columnwidth}{@{}Yrr@{}}\toprule
Tool library & Total (min) & Seconds/case\\\midrule
Qwen3.5-Plus: retrieval + repair & 48.15 & 2.626\\
GPT-5.3-Codex: retrieval + repair & 48.47 & 2.644\\
GPT-5.5: retrieval + repair & 47.17 & 2.573\\
GPT-5.5: direct generation & 47.28 & 2.579\\
\bottomrule\end{tabularx}\end{table}

\section{Case Study and Error Analysis}

\begin{table*}[!t]
    \caption{Representative cases of our method for medical calculation.}
    \label{tab:cases}
    \centering
    \scriptsize
    \renewcommand{\arraystretch}{1.02}
    \begin{tabularx}{\textwidth}{@{}p{0.13\textwidth}XXX@{}}
        \toprule
        & Case A & Case B & Case C\\
        \midrule
        Calculator
        & Creatinine Clearance, Cockcroft-Gault (lab test)
        & HEART Score (risk)
        & Steroid Conversion Calculator (dosage)\\
        Patient note (key facts)
        & 87-year-old man; height 163 cm; weight 48 kg; serum creatinine 1.4 mg/dL; the question asks Cockcroft-Gault clearance with weight selection.
        & 47-year-old man in the ED; diabetes, hypertension, atorvastatin use, normal ECG, troponin $>3\times$ upper limit, no prior atherosclerotic disease, slightly suspicious history.
        & Patient receives Hydrocortisone PO 162.935 mg; the question asks for the equivalent Dexamethasone PO dose, with the oral route preserved.\\
        Question
        & Creatinine clearance, mL/min, using adjusted body weight if it exceeds ideal.
        & HEART Score (0--10) at first admission, before treatment.
        & Equivalent Dexamethasone PO dose in mg.\\
        Generated tool pseudocode (written once by GPT-5.5)
        & \texttt{ibw = ideal\_weight(sex, height)}; \texttt{w = actual if actual < ibw else adjusted}; \texttt{crcl = ((140-age)*w)/(72*scr)}.
        & \texttt{risk\_count = diabetes + hypertension + hypercholesterolemia + ...}; \texttt{risk\_points = 2 if risk\_count >= 3 else 1 if risk\_count >= 1 else 0}.
        & \textcolor{red}{\texttt{if "cortisone" in name: return cortisone}}; \texttt{elif "hydrocortisone" in name: return hydrocortisone}; then convert by steroid-equivalence table.\\
        Extracted parameters (Qwen3-8B-SFT)
        & sex=Male, age=87 y, weight=48 kg, height=163 cm, creatinine=1.4 mg/dL.
        & age=47, hypertension=True, diabetes=True, \textcolor{red}{hypercholesterolemia=False}, ECG=Normal, troponin=$>3\times$, atherosclerotic=False, suspicion=Slightly suspicious.
        & input steroid=Hydrocortisone PO, dose=162.935 mg; target steroid=Dexamethasone PO.\\
        Tool computation
        & weight 48 $<$ ideal 59.6, so uses actual weight; $(140-87)\times48/(72\times1.4)=25.24$.
        & only \textcolor{red}{2 risk factors} counted, so risk block scores \textcolor{red}{1} and total = \textcolor{red}{4}.
        & Hydrocortisone is misread as cortisone: $162.935\times0.75/25=\textcolor{red}{4.88805}$ mg.\\
        Model answer
        & 25.24 mL/min
        & \textcolor{red}{4}
        & \textcolor{red}{4.89 mg}\\
        Gold answer
        & 25.2381
        & 5
        & 6.029 mg\\
        Cause
        & The validated tool executes the correct formula from correctly extracted inputs.
        & Extraction error: the extractor missed that atorvastatin implies hypercholesterolemia.
        & Generated-tool error: the normalizer matches \texttt{cortisone} before \texttt{hydrocortisone}.\\
        \bottomrule
    \end{tabularx}
\end{table*}

To characterize the remaining failure modes, we analyze the 146 incorrect final answers produced by the best configuration on the test set. We categorize the errors into two types. The \emph{Extraction error} means that at least one required calculator field extracted from the note mismatches the gold entity. The \emph{Generated-tool error} means that all required fields are correctly extracted, but the final result remains incorrect because of an error in the generated tool, the rounding rule, the unit convention, the date boundary, or the answer string, which still violates the benchmark contract.

Table~\ref{tab:cases} presents one successful case and the two representative error cases. In Case A, the Cockcroft-Gault calculation succeeds because all required variables are explicitly extracted and the tool computes the correct creatinine-clearance value. Case B illustrates an extraction error. The HEART Score tool correctly applies the scoring rule to the parameters it receives. However, the extraction model fails to infer hypercholesterolemia from the patient’s atorvastatin use. Case C illustrates a tool error. The extractor correctly identifies the input steroid, dose, route, and target steroid: Hydrocortisone PO, 162.935 mg, and Dexamethasone PO. The tool should convert the dose using the equivalence between oral hydrocortisone and oral dexamethasone, yielding the benchmark answer of 6.029 mg. However, the GPT-5.5-generated normalization logic searches for the substring cortisone before hydrocortisone, causing Hydrocortisone PO to be incorrectly normalized as cortisone. The resulting computation, $162.935\times0.75/25=4.88805$, is therefore based on the wrong source-steroid equivalence and produces an incorrect final answer of 4.89 mg. This error originates in the generated tool rather than in parameter extraction or answer formatting.

\begin{figure}[!htbp]
    \centering
    \includegraphics[width=0.96\columnwidth]{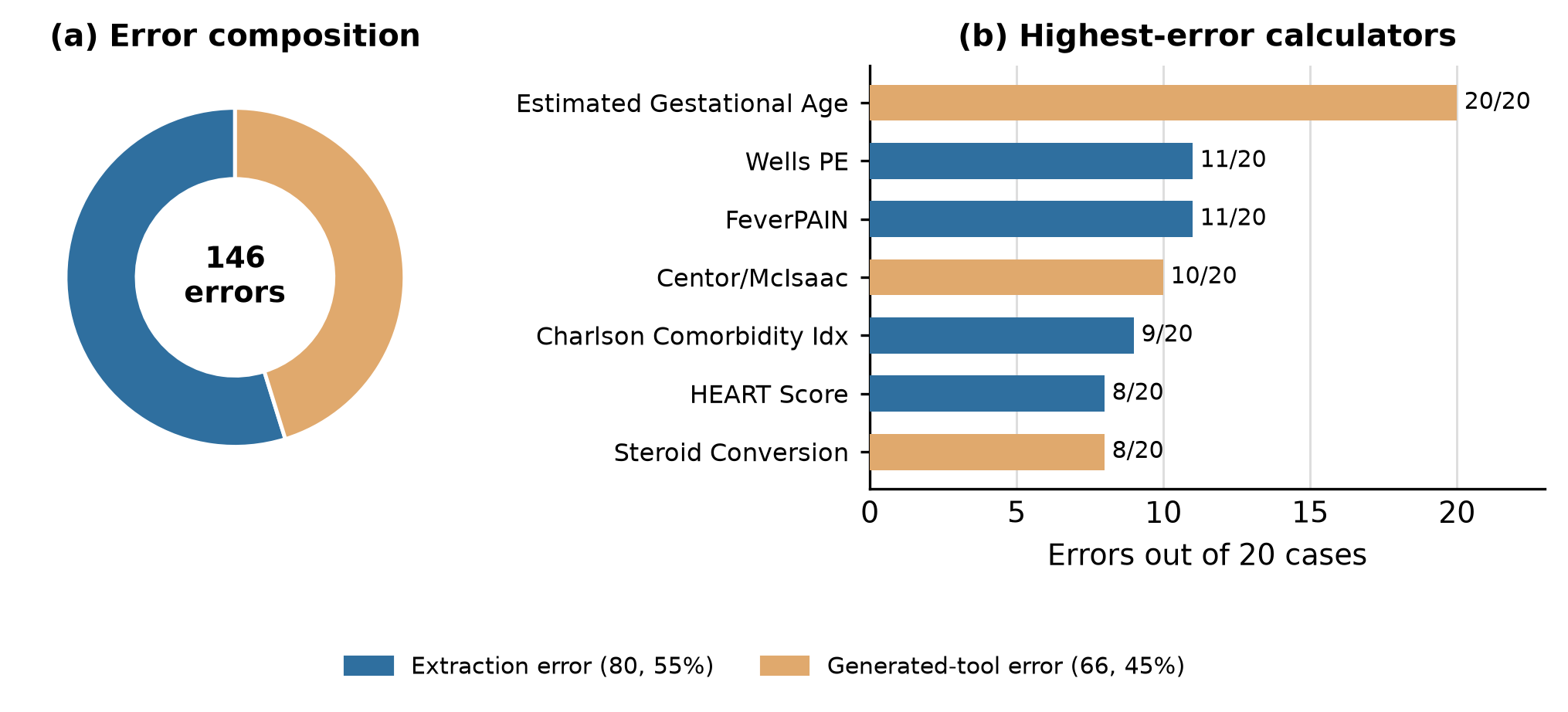}
    \caption{Error analysis of the best run. Extraction errors come from incorrectly extracted calculator fields, while Generated-tool errors occur after the required fields have been extracted correctly.}
    \label{fig:error}
\end{figure}

Figure~\ref{fig:error}(a) summarizes the overall error distribution: 80 errors result from extraction, whereas 66 result from the generated tool. Extraction errors are concentrated in calculators that require clinical-semantic interpretation, including diagnosis identification, temporal reasoning, implied comorbidities, and risk-factor assessment. Generated-tool errors occur more often when the necessary parameters are relatively easy to identify but correct evaluation depends on a specific formula branch, normalization rule, rounding procedure, unit convention, date boundary, or answer representation.

Figure~\ref{fig:error}(b) presents this distinction for the seven calculators with the largest numbers of errors. Each listed calculator contributes 20 test cases. Estimated Gestational Age is primarily affected by date-handling and output-contract errors, whereas Steroid Conversion is mainly affected by tool-rule errors, including normalization failures such as Case C. Wells PE, FeverPAIN, the Charlson Comorbidity Index, and the HEART Score are primarily affected by input-extraction errors because they require clinical judgments involving symptoms, timing, risk factors, and implied medical history.

\section{Conclusion}

This work presents a tool-augmented framework for reliable medical calculation from free-text patient notes. The framework separates calculator-rule retrieval, executable tool generation, schema-constrained parameter extraction, and deterministic tool execution. By grounding calculator tools in retrieved evidence and validating them through execution feedback, the approach reduces reliance on LLM formula recall and minimizes arithmetic errors. The separation between clinical interpretation and deterministic calculation also improves transparency. On the corrected MedCalc-Bench Verified test set, our method achieves the competitive performance, substantially outperforming prompting and SFT-based baselines. The analyses further show that reliable performance depends on both high-quality generated calculator tools and accurate schema-constrained extraction.

Despite these advantages, the study faces limitations that define its current scope. Extraction errors persist when the system must infer implicit information like medication-related conditions, negation, or temporality. Generated-tool errors arise when inputs are processed using incorrect normalization rules, formula branches, or output formats. Future work will address these challenges through reinforcement learning that rewards both accurate extraction of critical fields and correct final calculations. Furthermore, intelligent agents could improve reliability by verifying uncertain parameters, testing generated tools on boundary cases, and requesting clarification when key information is missing.

Evaluation concerns new patient notes for the same 55 calculators used during tool construction. Generalization to previously unseen calculators and clinical deployment remain untested.

\section*{Acknowledgment}
This work was supported by the National Natural Science Foundation of China (No. 62302076, 62276043, 62676065) and the Fundamental Research Funds for the Central Universities (No. DUT25YG108).

\bibliographystyle{IEEEtran}
\bibliography{references}

\end{document}